\documentclass[sigconf]{aamas}  % do not change this line!
\usepackage{booktabs}
\usepackage[utf8]{inputenc}
\usepackage{amsmath}

\usepackage{amsfonts}
\usepackage{mathtools}
\usepackage{graphicx}
\usepackage{float}
\usepackage{caption}
\usepackage{subcaption}
\usepackage{xcolor}
\usepackage{flushend}
\setcopyright{ifaamas}  % do not change this line!
\acmDOI{doi}  % do not change this line!
\acmISBN{}  % do not change this line!
\acmConference[ALA'20]
{Proc.\@ of the 19th International Conference on Autonomous Agents and Multiagent Systems (AAMAS 2020), B.~An, N.~Yorke-Smith, A.~El~Fallah~Seghrouchni, G.~Sukthankar (eds.)}{May 2020}{Auckland, New Zealand}  % do not change this line!
\acmYear{2020}  % do not change this line!
\copyrightyear{2020}  % do not change this line!
\acmPrice{}  % do not change this line!
\renewcommand\footnotetextcopyrightpermission[1]{} % removes footnote with conference information in first column
\newcommand*{\figuretitle}[1]{%
    {\centering%   <--------  will only affect the title because of the grouping (by the
    \textbf{#1}%              braces before \centering and behind \medskip). If you remove
    \par\medskip}%            these braces the whole body of a {figure} env will be centered.
}

\newcommand{\sa}{%
  \mathrel{%
    \vcenter{\offinterlineskip
      \ialign{##\cr$s_t = s,$\cr\noalign{\kern+1.5pt}$a_t = a$\cr}%
    }%
  }%
}

\begin{document}
\title{Useful Policy Invariant Shaping from Arbitrary Advice}
%\titlenote{Produces the permission block, and copyright information}

% AAMAS: as appropriate, uncomment one subtitle line; check the CFP
%\subtitle{Extended Abstract}
%\subtitle{Blue Sky Ideas Track}
%\subtitle{JAAMAS Track}
%\subtitle{Demonstration}
%\subtitle{Doctoral Consortium}

% AAMAS: submissions are anonymous for most tracks
\author{Paniz Behboudian}
\affiliation{\institution{University of Alberta}}
\email{behboudi@ualberta.ca}
\author{Yash Satsangi}
\affiliation{\institution{University of Alberta}}
\email{ysatsang@ualberta.ca}
\author{Matthew E. Taylor}
\affiliation{\institution{University of Alberta}}
\email{mtaylor3@ualberta.ca}
\author{Anna Harutyunyan}
\affiliation{\institution{Google, DeepMind}}
\email{harutyunyan@google.com}
\author{Michael Bowling}
\affiliation{\institution{University of Alberta}}
\email{mbowling@ualberta.ca}
% put your paper number here!

\begin{abstract}  % put your abstract here!
\textit{Reinforcement learning} (RL) is a powerful learning paradigm in which agents can learn to maximize sparse and delayed reward signals. Although RL has had many impressive successes in complex domains, learning can take hours, days, or even years of training data. A major challenge of contemporary RL research is to discover how to learn with less data.  Previous work has shown that domain information can be successfully used to shape the reward; by adding additional reward information, the agent can learn with much less data. Furthermore, if the reward is constructed from a potential function, the optimal policy is guaranteed to be unaltered. While such \emph{potential-based reward shaping} (PBRS) holds promise, it is limited by the need for a well-defined potential function. Ideally, we would like to be able to take arbitrary advice from a human or other agent and improve performance without affecting the optimal policy. The recently introduced \textit{dynamic potential based advice} (DPBA) method tackles this challenge by admitting arbitrary advice from a human or other agent and improves performance without affecting the optimal policy. The main contribution of this paper is to expose, theoretically and empirically, a flaw in DPBA. Alternatively, to achieve the ideal goals, we present a simple method called \textit{policy invariant explicit shaping} (PIES) and show theoretically and empirically that PIES succeeds where DPBA fails.

% This paper presents a simple method called policy invariant explicit shaping that can take arbitrary advise from a human or other agent and improve the performance of the agent without affecting the optimal policy. We expose a technical flaw in the recently introduced dynamic potential based advise (DPBA) method and show theoretically and empirically that PIES is a simple alternative that succeeds where DPBA fails

%In contrast, the recently-introduced \emph{dynamic PBRS} algorithm was designed to take arbitrary advice from a human or other agent and improve performance without affecting the optimal policy.
% We expose a technical flaw 
% In contrast, the recently-introduced \emph{dynamic PBRS} algorithm was designed to take arbitrary advice from a human or other agent and improve performance without affecting the optimal policy. The primary contribution of this paper is to expose, theoretically and empirically, a flaw in this well-cited, published approach. The secondary contribution is to introduce a simple alternative method that succeeds where dynamic PBRS can fail, successfully allowing arbitrary advice to improve learning speed without changing the optimal policy. 
\end{abstract}
  % put your semicolon-separated keywords here!

\maketitle
\section{Introduction}\label{sec:intro}

A \textit{reinforcement learning} (RL) agent aims to learn a policy (a mapping from states to actions) that maximizes a reward signal \cite{sutton1998reinforcement}. In many cases, the reward signal is sparse and delayed so that learning a good policy can take an excessively long time. For example, the Open AI Five agent \cite{OpenAI_dota} required 180 years worth of game experience per day of training; similarly, grand-master level StarCraft agent AlphaStar \cite{Vinyals2019}, required 16,000 matches as training data. One approach to accelerate learning is to add an external source of advice. The practice of providing an RL agent with additional rewards to improve learning is called \emph{reward shaping} and the additional reward is called the shaping reward. However, naively augmenting the original reward function with shaping can alter the optimal policy of the RL agent \cite{randlov1998learning}. For example, \citet{randlov1998learning} showed that adding a shaping reward (that at first seems reasonable) causes an agent learning how to ride a bicycle toward a goal to instead get ``distracted'' and ride in a loop, repeatedly collecting the shaping reward. 

\emph{Potential based reward shaping} (PBRS) \cite{ng1999policy,wiewiora2003principled,wiewiora2003potential} allows an RL agent to incorporate external advice without altering its optimal policy by deriving the shaping reward from a potential function.
Given a static potential function, PBRS defines the shaping reward as the difference in the potentials of states (or state-action pairs) when an agent makes a transition from one state to another. \citet{ng1999policy} showed that PBRS is guaranteed to be policy invariant: using PBRS does not alter the optimal policy. 

While PBRS achieves policy invariance, it may be difficult or impossible for a person or agent to express their advice as a potential-based function. Instead, it would be preferable to allow the use of more direct or intuitive advice in the form of an arbitrary function. The ideal reward shaping method then would have three properties:
\begin{enumerate}
\item Be able to use an arbitrary reward function as advice,
\item Keep the optimal policy unchanged in the presence of the additional advice, and
\item Improve the speed of learning of the RL agent when the advice is good.\footnote{We used ``good'' and ``bad'' in simple relative terms. We refer to advice as ``good'' to simply mean that one would expect that it would help the speed of learning, e.g., it rewards optimal actions more often than non-optimal actions.}
\end{enumerate}
\citet{harutyunyan2015expressing} attempted to tackle the same problem by proposing the framework of \textit{dynamic potential-based advice} (DPBA), where the idea is to dynamically learn a potential function from arbitrary advice, which can then be used to define the shaping reward. Importantly, the authors claimed that if the potential function is initialized to zero then DPBA is guaranteed to be policy invariant. We show in this work that this claim is not true, and hence, the approach is unfortunately {\em not} policy invariant. We confirm our finding theoretically and empirically. We then propose a fix to the method by deriving a correction term, and show that the result is theoretically sound, and empirically policy-invariant. However, our empirical analysis shows that the \textit{corrected} DPBA fails to accelerate the learning of an RL agent provided with useful advice.

We introduce a simple algorithm, \textit{policy invariant explicit shaping} (PIES), and show that PIES can allow for arbitrary advice, is policy invariant, and can accelerate the learning of an RL agent. PIES biases the agent's policy toward the advice at the start of the learning, when the agent is the most in need of guidance. Over time, PIES gradually decays this bias to zero, ensuring policy invariance. Several experiments confirm that PIES ensures convergence to the optimal policy when the advice is misleading and also accelerates learning when the advice is useful.

Concretely, this paper makes the following contributions:
\begin{enumerate}
    \item Identifies an important flaw in a published reward shaping method.
    \item Verifies the flaw exists, empirically and theoretically.  % by showing that the optimal policy can be altered by advice.
    \item Provides a correction to the method, but empirically shows that it introduces additional complications, where good advice no longer improves learning speed.
    \item Introduces and verifies a simple approach that achieves the goals of the original method. %verifying empirically and theoretically that it meets the three criteria desired from a reward shaping method.
\end{enumerate}

\section{Background}
\label{sec:background}
A \textit{Markov Decision Process} (MDP) \cite{puterman2014markov}, is described by the tuple $\langle S, A, T, \gamma, R \rangle$. At each time step, the environment is in a state $s \in S$, the agent takes an action $a \in A$, and the environment transitions to a new state $s' \in S$, according to the transition probabilities $T(s, a, s') = \Pr(s' | s, a)$. Additionally, the agent (at each time step) receives a reward for taking action $a$ in state $s$ according to the reward function $R(s,a)$. Finally, $\gamma$ is the discount factor, specifying how to trade off future rewards and current rewards.

A deterministic policy $\pi$ is a mapping from states to actions, $\pi : S \to A$, that is, for each state, $s$, $\pi(s)$ returns an action, $a = \pi(s)$. The \emph{state-action value function} $Q^{\pi}(s,a)$ is defined as the expected sum of discounted rewards the agent will get if it takes action $a$ in state $s$ and follows the policy $\pi$ thereafter.
\begin{equation*}
Q^{\pi}(s,a) = \mathbb{E}\left[\sum_{k=0}^{\infty} \gamma^{k} R(s_{t+k},a_{t+k}) \middle| s_t = s, a_t = a, \pi \right]. 
\end{equation*}
%\MET{I'd only use numbered equations where you have to refer to them later. If you don't need to refer to them, they don't need a number.}
The agent aims to find the optimal policy denoted by $\pi^*$ that maximizes the expected sum of discounted rewards, and the state-action value function associated with $\pi^*$ is called the optimal state-action value function, denoted by $Q^{*}(s,a)$:
\begin{equation*}
Q^*(s,a) = \max_{\pi \in \prod} Q^{\pi}(s,a), 
\end{equation*}
where $\prod$ is the space of all policies.

The action value function for a given policy $\pi$ satisfies the \emph{Bellman equation}: 
\begin{equation*}
Q^{\pi}(s,a) = R(s,a) + \gamma \mathbb{E}_{s',a'}[Q^{\pi}(s',a')],  
\end{equation*}
where $s'$ is the state at the next time step and $a'$ is the action the agent takes on the next time step, and this is true for all policies.

The Bellman equation for the optimal policy $\pi^*$ is called the \emph{Bellman optimality equation}:
\begin{equation*}
Q^{*}(s,a) = R(s,a) + \gamma \mathbb{E}_{s',a'}[Q^{*}(s',a')].
\end{equation*}
Given the optimal value function $Q^{*}(s,a)$, the agent can retrieve the optimal policy by acting greedily with respect to the optimal value function:
\begin{equation*}
\pi^*(s) = \arg\max_{a \in A} Q^{*}(s,a).
\end{equation*}

The idea behind many reinforcement learning algorithms is to learn the optimal value function $Q^*$ iteratively. For example, Sarsa~\cite{sutton1998reinforcement} learns $Q$-values with the following update rule, at each time step $t$ ($Q_0$ can be initialized arbitrarily): 
\begin{equation}
Q_{t+1}(s_t,a_t) = Q_{t}(s_t,a_t) + \alpha_t \delta_{t},
\label{eq:updaterule}
\end{equation}
where 
\begin{equation*} 
\delta_{t} = R(s_{t}, a_{t}) + \gamma Q_{t}(s_{t+1}, a_{t+1}) - Q_{t}(s_t, a_t)
\end{equation*}
is the \emph{temporal-difference} error (TD-error), $s_t$ and $a_t$ denotes the state and action at time step $t$, $Q_{t}$ denotes the estimate of $Q^*$ at time step $t$, and $\alpha_t$ is the learning rate at time step $t$. Under certain conditions, these $Q$ estimates are guaranteed to converge to $Q^*$ for all $s,a$, and the policy converges to $\pi^*$~\cite{sutton1998reinforcement}.
% \ys{Decide later if we need $s_t$ notation or $s$ and $s'$ is enough.}
%\resizebox{0.5\textwidth}{!}{$Q_{t+1}(s_t,a_t) = Q_{t}(s_t,a_t) + \alpha[R(s_{t+1}, a_{t+1}) + \gamma Q_{t}(s_{t+1}, a_{t+1}) - Q_{t}(s_t, a_t)],$}

\subsection{Potential-Based Reward Shaping}
\label{sec:PBRS}

%and learning the value function is hard
In cases where the rewards are sparse, reward shaping can help the agent learn faster by providing an additional \emph{shaping reward} 
% $F: S \times A \times S \to \mathbb{R}$
$F$. However, the addition of an arbitrary reward can alter the optimal policy of a given MDP \cite{randlov1998learning}. 
% \MET{I'd cite the Randløv and Alstrøm example of the bicycle}
% \subsection{Potential-Based Reward Shaping}
\textit{Potential-based reward shaping} (PBRS) addresses the problem of adding a shaping reward function $F$ to an existing MDP reward function $R$, without changing the optimal policy by defining $F$ to be the difference in the \emph{potential} of the current state $s$ and the next state $s'$ \cite{ng1999policy}. Specifically, PBRS restricts the shaping reward to the following form: $F(s, s') \coloneqq \gamma \Phi(s') - \Phi(s)$, where $\Phi: S \to \mathbb{R}$ is the potential function. \citet{ng1999policy} showed that expressing $F$ as the difference of potentials is the sufficient condition for the agent to be \emph{policy invariant}. That is, if the original MDP $\langle S, A, T, \gamma, R \rangle$ is denoted by $M$ and the shaped MDP $\langle S, A, T, \gamma, R + F \rangle$ is denoted by $M'$ ($M'$ is same as $M$ but offers the agent an extra reward $F$ in addition to $R$) then the optimal value function of $M$ and $M'$ for any state-action pair $(s,a)$ satisfy:
\begin{equation*}
Q^{*}_{M'}(s,a) = Q^{*}_{M}(s,a) - \Phi(s) \,
\end{equation*}
where $\Phi$ is the \textit{bias term}. 
% \MET{I'm confused - why does the previous equation need to reason over over all states and actions to compute a single Q(s,a)?}% needed to recover the optimal policy.
Given $Q_{M'}^{*}$, the optimal policy $\pi^{*}$ can simply be obtained by adding the bias term as:
\begin{equation*}
    \pi^{*}(s) = \arg\max_{a \in A} Q^{*}_{M}(s,a) = \arg\max_{a \in A} (Q^{*}_{M'}(s,a) + \Phi(s)). 
\end{equation*}

Because the bias term only depends on the agent's state, the optimal policy of the shaped MDP $M'$, does not differ from that of the original MDP $M$. To also include the shaping reward on actions, Wiewiora et al.~\cite{wiewiora2003principled} extended the definition of $F$ to state-action pairs by defining $F$ to be: $F(s, a, s', a') \coloneqq \gamma \Phi(s', a') - \Phi(s, a)$, where $\Phi$ is dependent on both the state and the action of the agent. Now the bias term is also dependent on the action taken at state $s$, therefore the agent must follow the policy 
\begin{equation*}
\pi^*(s) = \arg\max_{a \in A} (Q^*_{M'}(s,a) + \Phi(s,a))
\end{equation*}
in order to be policy-invariant.

\subsection{Dynamic Potential-Based Shaping}
\label{sec:dynamic_PBRS}

PBRS, as discussed in Section~\ref{sec:PBRS}, is restricted to external advice that can be expressed in terms of a potential function. Therefore, it does not satisfy the first of the three goals; i.e., it cannot admit arbitrary advice. Finding a potential function $\Phi$ that accurately captures the advice can be challenging. To allow an expert to specify an arbitrary function $R^{\text{\it{expert}}}$ and still maintain all the properties of PBRS one might consider \textit{dynamic PBRS}. 

Dynamic PBRS uses a potential function $\Phi_t$ that can be altered online to form the dynamic shaping reward $F_t$, where subscript $t$ indicates the time over which $F$ and $\Phi$ vary. \citet{devlin2012dynamic} used dynamic PBRS as $F_{t+1}(s,s') := \gamma \Phi_{t+1}(s') - \Phi_t(s)$, where $t$ and $t+1$ are the times that the agent arrives at states $s$ and $s'$, respectively. They derived the same guarantees of policy invariance for dynamic PBRS as static PBRS. 
To admit an arbitrary reward, \citet{harutyunyan2015expressing} suggested learning a dynamic potential function $\Phi_t$ given the external advice in the form of an arbitrary bounded function, $R^{\text{\it{expert}}}$. 
To do so, the following method named \textit{dynamic potential based advice} (DPBA) is proposed: define $R^{\Phi} \coloneqq -R^{\text{\it{expert}}}$, and learn a \emph{secondary value function} $\Phi$ via the following update rule
% (after zero initialization of $\Phi_{0}$)
at each time step:
\begin{equation}\label{eq:phi_update_rule}
\Phi_{t+1}(s, a) \coloneqq \Phi_t(s, a) + \beta \delta^{\Phi}_t
\end{equation}
% $$\Phi_{t+1}(s, a) \coloneqq \Phi_t(s, a) + \beta \delta^{\Phi}_t $$
 where $\Phi_t(s,a)$ is the current estimate of $\Phi$, $\beta$ is the $\Phi$ function's learning rate, and
 $$\delta^{\Phi}_t \coloneqq R^{\Phi}(s, a) + \gamma \Phi_{t+1}(s',a') - \Phi_t(s, a)$$
 is the $\Phi$ function's TD error. All the while, the agent learns the $Q$ values using Sarsa (i.e., according to Equation \ref{eq:updaterule}). In addition to the original reward $R(s,a)$ the agent receives a shaping reward given as:
 \begin{equation}\label{eq:dynamicf}
F_{t+1}(s,a,s',a') \coloneqq \gamma\Phi_{t+1}(s', a') - \Phi_t(s, a),
 \end{equation}
that is, the difference between the consecutively updated values of $\Phi$. 

~\citet{harutyunyan2015expressing} suggested that with this form of reward shaping, $Q^*_M(s,a) = Q^*_{M'}(s,a) + \Phi_0(s,a)$ for every $s$ and $a$ and therefore to obtain the optimal policy $\pi^*$, the agent should be greedy with respect to $Q^*_{M'}(s,a) + \Phi_0(s,a)$ by the following rule: 
\begin{equation} \label{eq:policy_uncorrected}
\pi^*(s) = \arg\max_{a \in A} (Q^*_{M'}(s,a) + \Phi_{0}(s,a)),
\end{equation}
and thus if $\Phi_{0}(s,a)$ is initialized to zero, the above biased policy in Equation \ref{eq:policy_uncorrected}, reduces to the original greedy policy: 
\begin{equation*}
\pi^*(s) = \arg\max_{a \in A} Q^*_{M'}(s,a) = \arg\max_{a \in A} Q^*_{M}(s,a).
\end{equation*}
DPBA was empirically evaluated on two episodic tasks: a $20\times20$ grid-world and a cart-pole problem. In the grid-world experiment, the agent starts each episode from the top left corner until it reaches the goal located at the bottom right corner, within a maximum of 10,000 steps. The agent can move along the four cardinal directions and the state is the agent's coordinates $(x,y)$. The reward function is +1 upon arrival at the goal state and 0 elsewhere. The advice, $R^{\text{\it{expert}}}$, for any state action is:
$$
R^{\text{\it{expert}}}(s,a) := \begin{cases}
+1 & \textrm{\textit{if a is} right \textit{or} down} \\
0 & \text{\it{otherwise}}
\end{cases}.$$
This paper replicates the same grid-world environment in our later experiments.

In the cart-pole task \cite{michie1968boxes}, the goal is to balance a pole on top of a cart as long as possible. The cart can move along a track and each episode starts in the middle of the track with the pole upright. There are two possible actions: applying a force of +1 or -1 to the cart. The state consists of a four dimensional continuous vector, indicating the angle and the angular velocity of the pole, and the position and the velocity of the cart. An episode ends when the pole has been balanced for 200 steps or the pole falls, and the reward function encourages the agent to balance the pole.
%, or the cart move beyond the track's boundaries. 
To replicate this experiment\footnote{To assist with reproducibility, after acceptance we will release our code necessary to run all experiments in this paper.}, this paper used the OpenAI Gym \cite{brockman2016openai} implementation (cartpole-v0).\footnote{There are slight differences between our implementation of cart-pole and the version used in the DPBA paper \cite{harutyunyan2015expressing}, making the results not directly comparable. In that paper, 1) if the cart attempts to move beyond the ends of the track, the cart bounces back, and 2) there is a negative reward if the pole drops and otherwise the reward is zero. In contrast, in OpenAI Gym, 1) if the cart moves beyond the track's boundaries, the episode terminates, and 2) the reward function is +1 on every step the pole is balanced and 0 if the pole falls.} The advice for this task is defined as:
$$
R^{\text{\it{expert}}}(s,a) := o(s,a) \times c,
$$
where $o:S \times A \to \{0,1\}$ is a function that triggers when the pole direction is aligned with the force applied to the cart (i.e., when the cart moves in the same direction as the pole is leaning towards, the agent is rewarded). We set $c=0.1$.
\begin{figure}
\figuretitle{DPBA}
    \centering
    \begin{subfigure}{0.4\textwidth}
        \centering
        \includegraphics[width=\textwidth]{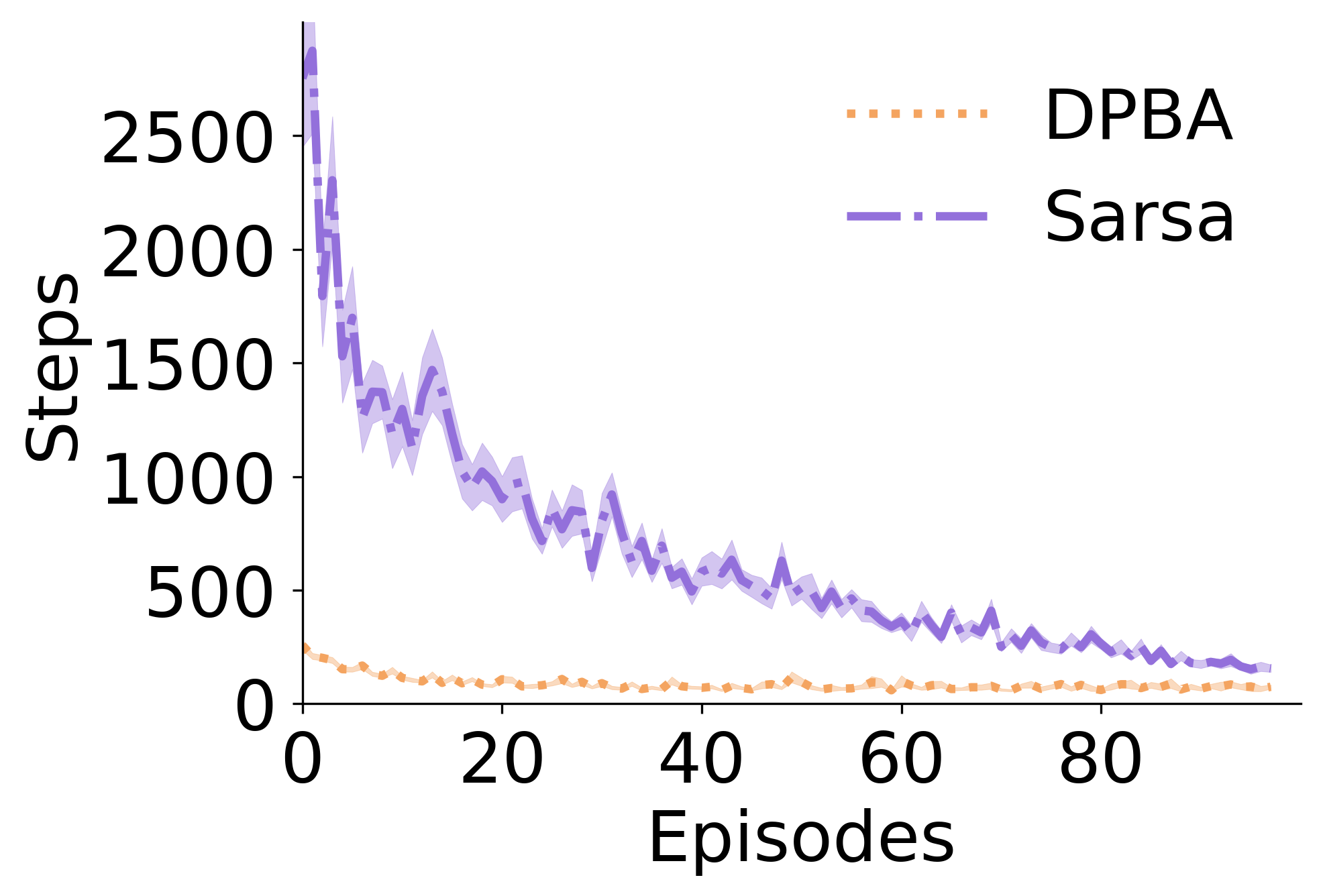}
        \caption{Grid-World}
        \label{fig:pbrs_gw}
    \end{subfigure}
    \begin{subfigure}{0.4\textwidth}
        \centering
    \includegraphics[width=\textwidth]{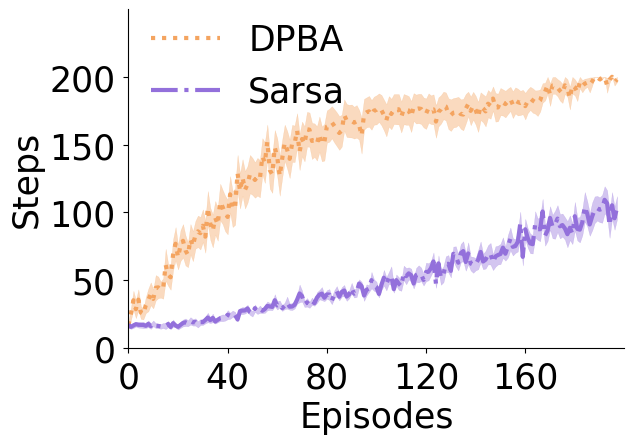}
        \caption{Cart-Pole}
        \label{fig:pbrs_cp}
    \end{subfigure}
    \caption{The y-axis shows the number of time steps taken to finish each episode (on x-axis) averaged over (a) 50  and (b) 30 runs. The shaped agent with DPBA is compared with a Sarsa learner without shaping in a) grid-world and b) cart-pole domains. Shaded areas correspond to the standard error.}
    \label{fig:pbrs}
\end{figure}
\begin{table}
  \caption{Parameters values for Figure \ref{fig:pbrs} (b)}
  \label{table:pbrs_cp}
  \begin{tabular}{cccl}
    \toprule
    Agent&$\alpha$&$\beta$\\
    \midrule
    Sarsa & 0.1 & -\\
    DPBA & 0.02 & 0.1\\
  \bottomrule
\end{tabular}
\end{table}

 Figure \ref{fig:pbrs} shows the performance of the DPBA method, compared to a simple Sarsa learner not receiving any expert advice, in the grid-world and the cart-pole domains. We used the same set of hyper-parameters as used in \cite{harutyunyan2015expressing} for the grid-world. For learning the cart-pole task, the agent used a linear function approximation for estimating the value function via Sarsa($\lambda$) and tile-coded feature representation \cite{sutton1996generalization} with the implementation from the open-source software (publicly available at Richard Suttons's website). The weights for $Q$ and $\Phi$ were initialized uniformly random between 0 and 0.001. For tile-coding, we used 8 tilings, each with $2^4$ tiles (2 for each dimension). We used a wrapping tile for the angle of the pole for a more accurate state-representation. With a wrapping tile one can generalize over a range (e.g. $[0,2\pi]$) rather than stretching the tile to infinity, and then wrap-around. $\lambda$ was set to 0.9 and $\gamma$ to 1. For learning rates of $Q$ and $\Phi$ value functions, $\alpha$ and $\beta$, we swept over the values $[0.001, 0.002, 0.01, 0.1, 0.2]$. The best parameter values according to the area under curve (AUC) of each line for Figure \ref{fig:pbrs} (b) is reported in Table \ref{table:pbrs_cp}.

These results agree with the prior work, showing that the agent using the DPBA method learned faster with this good advice, relative to not using the advice (i.e. the DPBA line is converging faster to the optimal behaviour). Note that in the grid-world task the desired behaviour is to reach the goal as fast as possible. Consequently, for this task the lower is better in plots (such as the ones in Figure \ref{fig:pbrs}) showing steps (y-axis) versus episodes (x-axis). In contrast, the nature of the goal in the cart-pole task dictates to take as much steps as possible during each episode, as the desired behaviour. Therefore, for cart-pole higher is better in the same plot with the same axes. The results in Figure \ref{fig:pbrs} show that DPBA method satisfies criterion 1 (it can use arbitrary rewards) and criterion 3 (good advice can improve performance). However, as we argue in the next section, a flaw in the proof of the original paper means that criterion 2 is not satisfied: the optimal policy can change, i.e., advice can cause the agent to converge to a sub-optimal policy. This was not empirically tested in the original paper and thus this failure was not noticed.

\section{DPBA Can Affect the Optimal Policy}
The previous section described DPBA, a method that can incorporate an arbitrary expert's advice in a reinforcement learning framework by learning a potential function $\Phi$ iteratively and concurrently with the shaped state-action values, $Q_{M'}$. \citet{harutyunyan2015expressing} claimed that if the initial values of $\Phi$, $\Phi_{0}$, are initialized to zero, then the agent can simply follow a policy that is greedy with respect to $Q_{M'}$ to achieve policy invariance. In this section we show that this claim is unfortunately not true: initializing $\Phi_{0}(s,a)$ to zero is not sufficient to guarantee policy invariance. 

To prove our claim, we start by defining terms. We will compare Q-value estimates for a given policy $\pi$ in two MDPs, the original MDP denoted by $M$ described by the tuple $\langle S, A, T, \gamma, R \rangle$, and the MDP that is shaped by DPBA, $M'$, described by the tuple $\langle S, A, T, \gamma, R + F_{t+1} \rangle$, where $F_{t+1}(s,a,s',a') = \gamma \Phi_{t+1}(s',a') - \Phi_{t}(s,a)$.

Let $R'_{t+1} := R + F_{t+1}$, given a policy $\pi$, at any time step $t$, $Q^{\pi}_{M'}$ can be defined as:
\begin{align}
    Q^{\pi}_{M'}(s,a) &= \mathbb{E} \left[ \sum_{k=0}^{\infty} \gamma^{k}R'_{t+k+1}(s_{t+k}, a_{t+k}, s_{t+k+1}, a_{t+k+1}) \middle| \sa \right]. \nonumber 
\end{align}
By writing $R'$ in terms of $R$ and $F$:
\begin{align*}
=\mathbb{E}&\left[\sum_{k=0}^{\infty}\gamma^{k} \left(R(s_{t+k}, a_{t+k}) \right.\right.\\
&\qquad\left.\left.+ F_{t+k+1}(s_{t+k}, a_{t+k}, s_{t+k+1}, a_{t+k+1})\right) \middle|
\vphantom{\sum_{k=0}^{\infty}} \sa \right]
\end{align*}
The first term in the above expression (after separating the expectation) is the value function for the original MDP $M$ for policy $\pi$.
\begin{align}
=&\underbrace{\mathbb{E}\left[\sum_{k=0}^{\infty}\gamma^{k} R(s_{t+k}, a_{t+k})\middle| \sa \right]}_{=Q^\pi_M(s,a)}\nonumber\\
&+\mathbb{E}\left[\sum_{k=0}^{\infty}\gamma^{k} F_{t+k+1}(s_{t+k}, a_{t+k}, s_{t+k+1}, a_{t+k+1}) \middle| \sa \right]\label{eq:second}
\end{align}
The second term in Equation \ref{eq:second} can be expanded based on Equation \ref{eq:dynamicf} as follows:
\begin{align}
    \mathbb{E}&\left[\sum_{k=0}^{\infty}\gamma^{k} F_{t+k+1}(s_{t+k}, a_{t+k}, s_{t+k+1}, a_{t+k+1}) \middle| \sa \right] \nonumber \\
     &= \mathbb{E}\left[\sum_{k=0}^{\infty}\left(\gamma^{k+1} \Phi_{t+k+1}(s_{t+k+1}, a_{t+k+1}) \right. \right. \nonumber\\
     &\left.\left. \qquad \qquad - \gamma ^{k}\Phi_{t+k}(s_{t+k}, a_{t+k})\right) \middle| \sa  \vphantom{\sum_{k=0}^{\infty}}\right] \label{eq:third}
\end{align}
The two terms inside the infinite summation look quite similar, motivating us to rewrite one of them by shifting its summation variable $k$. This shift will let identical terms be cancelled out. However, we need to be careful. First, we rewrite the sums in their limit form. An infinite sum can be written as:
\begin{equation*}\label{eq:infsum}
    \sum_{i=i_0}^{\infty}x_i \coloneqq \lim_{W \to\infty}\sum_{i=i_0}^{W}x_i.
\end{equation*}
Using this definition, in Equation \ref{eq:third} we can shift the first term's iteration variable as:
\begin{align}
=&\mathbb{E}\left[\lim_{W \to \infty}\left(\sum_{k=1}^{W+1}\gamma^{k}\Phi_{t+k}(s_{t+k}, a_{t+k}) \right. \right. \nonumber\\
&\left.\vphantom{\sum_{k=1}^{W+1}}\qquad\qquad - \sum_{k=0}^{W}\gamma^{k} \Phi_{t+k}(s_{t+k}, a_{t+k})\left.\vphantom{\sum_{k=1}^{W+1}}\right) \middle| \sa \right]\nonumber\\
=&\mathbb{E}\left[\vphantom{\left(\lim_{W\to\infty}\gamma^{W+1} \Phi_{t+W+1}(s_{t+W+1},a_{t+W+1})\right)}\lim_{W\to\infty} \right. \left(\gamma^{W+1} \Phi_{t+W+1}(s_{t+W+1}, a_{t+W+1}) \right. \nonumber\\
&\qquad\qquad- \left. \left. \gamma ^{0}\Phi_t(s_t, a_{t})\right)\middle| \right.
\left.\sa\vphantom{\left(\lim_{W\to\infty}\gamma^{W+1} \Phi_{t+W+1}(s_{t+W+1},a_{t+W+1})\right)}\right]\label{eq:f2}
\end{align}
% \begin{align}
% =&\mathbb{E}\left[\lim_{W \to \infty}\left(\sum_{k=1}^{W+1}\gamma^{k}\Phi_{t+k}(s_{t+k}, a_{t+k}) \right. \right. \nonumber\\
% &\left.\vphantom{\sum_{k=1}^{W+1}}\qquad\qquad - \sum_{k=0}^{W}\gamma^{k} \Phi_{t+k}(s_{t+k}, a_{t+k})\left.\vphantom{\sum_{k=1}^{W+1}}\right) \middle| \sa \right]\nonumber\\
% =&\mathbb{E}\left[\lim_{W\to\infty}\left(\underbrace{\gamma^{W+1} \Phi_{t+W+1}(s_{t+W+1}, a_{t+W+1})}_{\text{\romannumeral1}} - \underbrace{\gamma ^{0}\Phi_t(s_t, a_{t})}_{\text{\romannumeral2}} \right)\middle| \right.\nonumber\\
% &\left.\qquad\sa\vphantom{\left(\lim_{W\to\infty}\underbrace{\gamma^{W+1} \Phi_{t+W+1}(s_{t+W+1},a_{t+W+1})}_{\text{\romannumeral1}}\right)}\right]\label{eq:f2}
% \end{align}
In Equation \ref{eq:f2}, if $\Phi_t^{\pi}(s, a)$ is bounded, then the first term inside the limit will go to 0 as $W$ approaches infinity\footnote{Note that this term will go to zero only for infinite horizon MDPs. In practice, it is common to assume a finite horizon MDP with a terminal state, in such cases, this extra term will remain and must be removed, for example, by defining the potential of the terminal state to be zero.}, and the second term does not depend on $W$ and can be pulled outside the limit:
\begin{align}
&\mathbb{E}\left[\sum_{k=0}^{\infty}\gamma^{k} F_{t+k+1}(s_{t+k}, a_{t+k}, s_{t+k+1}, a_{t+k+1}) \middle| \sa \right] \nonumber \\
&=\mathbb{E}\left[-\Phi_t(s_t, a_{t}) \middle| \sa \right] = -\Phi_t(s,a) \label{eq:f3}
\end{align}
Back to Equation \ref{eq:second}, if we apply Equation \ref{eq:f3}, we will have:
\begin{equation*}
 Q^{\pi}_{M'}(s,a) = Q^{\pi}_M(s,a) - \Phi_t(s,a),
 \label{eq:q_correct}
\end{equation*}
for all $\pi$ (given that $s_t= s$ and $a_t = a$).
Thus, for an agent to retrieve the optimal policy $\pi^*_{M}$ given $Q^{*}_{M'}(s,a)$, it must act greedily with respect to $Q^{*}_{M'}(s,a) + \Phi_t(s,a)$:
\begin{equation}\label{eq:policy_corrected}
\pi^{*}_{M}(s) = \arg\max_{a \in A} \left(Q^{*}_{M'}(s,a) + \Phi_t(s,a)\right).
\end{equation}

% \ys{maybe it is an option to device a notation for conditioning on $s_t = s, a_t = a$, something like $ | H_t$ denotes that $ | s_t = s, a_t = a$.}

Equation \ref{eq:policy_corrected} differs from Equation \ref{eq:policy_uncorrected} (corresponding to Equation 17 in \citet{harutyunyan2015expressing}) in that the bias term is $\Phi_{t}$ and not $\Phi_{0}$. In other words, at every time step the $Q$ values of the agent are biased by the \emph{current estimate} of the potential function and not by the initial value of the potential function. The derived relation in Equation 17 of \citet{harutyunyan2015expressing} is only valid for the first state-action pair that the agent visits (at $t=0$). For the rest of the time steps, it is not accurate to use the first time step's bias term to compensate the shaped value function. Thus, the zero initialization of $\Phi$ is not a sufficient condition for the agent to recover the true $Q$ values of the original MDP, and it cannot be used to retrieve the optimal policy of the original MDP.

\subsection{Empirical Validation: Unhelpful Advice}
\label{sec:corrected_DPBA_bad_advice}
\begin{figure}[ht!]
\figuretitle{Toy Example Domain}
    \centering
    \begin{subfigure}{0.18\textwidth}
        \centering
        \includegraphics[width=\textwidth]{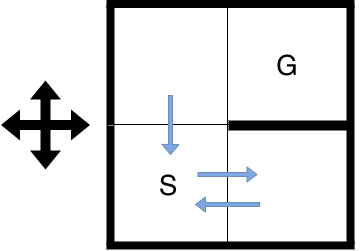}
        \caption{Bad expert}
        \label{fig:toy_example_bad}
    \end{subfigure}
    \begin{subfigure}{0.18\textwidth}
        \centering
        \includegraphics[width=2.3cm]{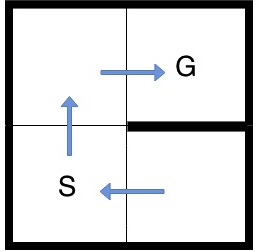}
        \caption{Good expert}
        \label{fig:toy_example_good}
    \end{subfigure}
    \caption{The toy example domain with advised transitions indicated by blue arrows. The bad expert in (a) tries to keep the agent away from the goal while the good expert in (b) rewards transitions towards the goal.}
    \label{fig:toy_example}
\end{figure}

We empirically validate the result above with a set of experiments. First, consider a deterministic $2\times2$ grid-world which we refer to it as the \textit{toy example}, depicted in Figure \ref{fig:toy_example}. The agent starts each episode from state S and can move in the four cardinal directions (as depicted in the figure) until it reaches the goal state G (within a maximum of 100 steps). Moving towards a wall (indicated by bold lines), causes no change in the agent's position. The reward is 0 on every transition except the one ending in the goal state, resulting in a reward of +1 and episode termination. 
For advice, we assume that the ``expert'' rewards the agent for making transitions away from the goal. Blue arrows inside the grid in Figure \ref{fig:toy_example} (a) represent the expert advised state-transitions. 
%The adversarially advised transitions are intended to keep the agent away from the goal by rewarding it to stay in a loop. 
The agent receives a $+1$ from the expert by executing the advised transitions. Because this advice is encouraging poor behavior, we expect that it would slow down the learning (rather than accelerate it), but if a shaping method is policy invariant, the agent should still eventually converge to the optimal policy. 
% \ys{One thing I felt from the presentation and from reader point of view is that the reviewer might doubt that the agent is not converging because of low value of exploration factor epsilon. Maybe it is useful to run the same experiment with a couple more values of epsilon.}\pb{The epsilon was initialized to 0.1 and decreased by $\frac{1}{300}$ at the end of each episode till episode 300.}

The learner uses Sarsa(0) to estimate $Q$ values with $\gamma=0.3$. We ran the experiment for both the learner with \textit{corrected DPBA}, Equation \ref{eq:policy_corrected}, and the learner with \textit{DPBA}, Equation \ref{eq:policy_uncorrected}, using an $\epsilon$-greedy policy. $\Phi$ and $Q$ were initialized to 0 and $\epsilon$ was decayed from 0.1 to 0. For the learning rates of the $Q$ and $\Phi$ value functions, $\alpha$ and $\beta$, we swept over the values $[0.05, 0.1, 0.2, 0.5]$. The best values, according to the AUC of the each line are reported in Table \ref{table:pbrs_toy_example_params}.

\begin{figure}[ht!]
\figuretitle{Corrected DPBA with Bad Advice}
    \centering
        \includegraphics[width=0.45\textwidth]{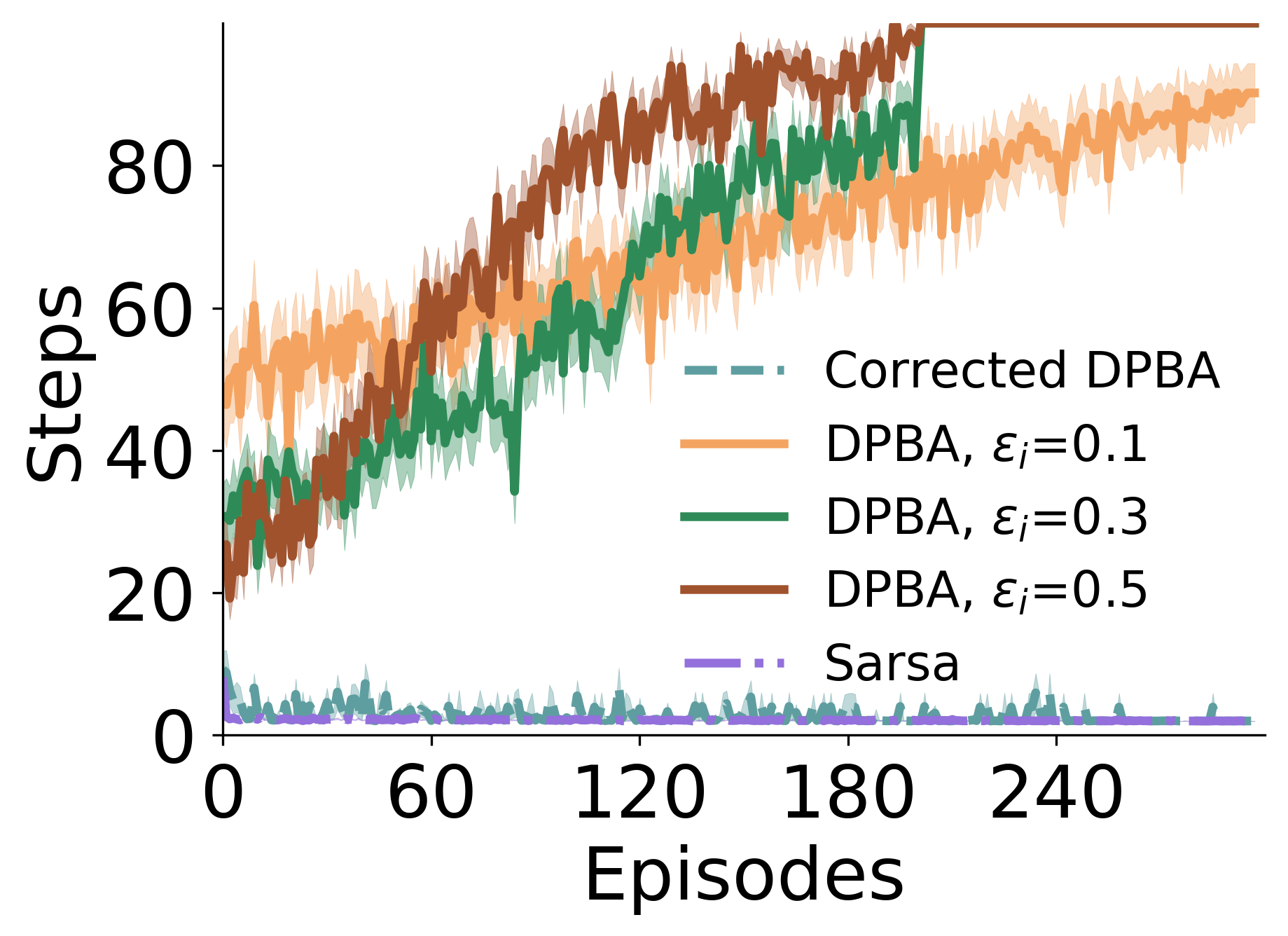}
    % \begin{subfigure}{0.4\textwidth}
    %     \centering
        \caption{The y-axis shows number of time steps taken to finish each episode with the bad expert. The shaped agent with corrected DPBA is compared with the shaped agent with DPBA and an unshaped Sarsa agent. Shaded areas correspond to the standard error averaged over 50 runs.}
    \label{fig:toy_example_corrected_bad}
\end{figure}

% \begin{table}[ht!]
%     \centering
%     \caption{Parameter values for Figures \ref{fig:toy_example_corrected_bad} and \ref{fig:toy_example_corrected_good}}
%     \label{table:pbrs_toy_example_params}
%     \begin{tabular}{c|cc|}
%         \cline{2-3}
%          & $\alpha$ & $\beta$ \\ \hline
%         \multicolumn{1}{|c|}{Sarsa} & 0.05 & - \\ \hline
%         \multicolumn{1}{|c|}{DPBA, good advice} & 0.2 & 0.5 \\ \hline
%         \multicolumn{1}{|c|}{DPBA, bad advice} & 0.2 & 0.5 \\ \hline
%         \multicolumn{1}{|c|}{corrected DPBA, good advice} & 0.2 & 0.1 \\ \hline
%         \multicolumn{1}{|c|}{corrected DPBA, bad advice} & 0.05 & 0.2 \\ \hline
%     \end{tabular}
% \end{table}
\begin{table}[ht!]
    \caption{Parameters values for Figures \ref{fig:toy_example_corrected_bad} and \ref{fig:toy_example_corrected_good}}
    \label{table:pbrs_toy_example_params}
  \begin{tabular}{ccl}
    \toprule
    Agent&$\alpha$&$\beta$\\
    \midrule
    Sarsa & 0.05 & - \\
    DPBA, good advice & 0.2 & 0.5\\
    DPBA, bad advice & 0.2 & 0.5\\
    corrected DPBA, good advice & 0.2 & 0.1\\
    corrected DPBA, bad advice & 0.05 & 0.2\\
  \bottomrule
\end{tabular}
\end{table}
Figure~\ref{fig:toy_example_corrected_bad} depicts the length of each episode as the number of steps taken to finish the episode. The \textit{Sarsa} line indicates the learning curve for a Sarsa(0) agent without shaping. Figure~\ref{fig:toy_example_corrected_bad} shows that DPBA does not converge to the optimal policy with a bad advice. This figure also confirms our result that $\Phi_t$ (and not $\Phi_0$ as proposed in \citet{harutyunyan2015expressing}) is the sufficient correction term to recover the optimal policy for maximizing the MDP's original reward.
% refuting the claim that DPRBS is policy invariant. 
% Initializing $\Phi_{0}$ to zero as proposed by \citet{harutyunyan2015expressing} is not a sufficient condition to recover the optimal policy for maximizing the MDP's original reward.
Finally, we verify that this is not simply an artefact of the agent exploiting too soon, and repeat the same experiments for different exploration rates, $\epsilon$. We considered two more different values for initial exploration rate, $\epsilon_i$: 0.3 and 0.5. The corresponding lines in Figure \ref{fig:toy_example_corrected_bad} confirm additional exploration does not let DPBA obtain the optimal policy. Figure \ref{fig:toy_example_corrected_bad} also confirms that the corrected policy as derived in Equation \ref{eq:policy_corrected} converges to the optimal policy even when the expert advice is bad.

\subsection{Empirical Validation: Helpful Advice}
\label{sec:corrected_DPBA_good_advice}
The previous section showed that DPBA is not a policy invariant shaping method since initializing the values of $\Phi$ to zero is not a sufficient condition for policy invariance. We showed that DPBA can be corrected by adding the correct bias term and indeed policy invariant. While the addition of the correct bias term guarantees policy invariance, we still need to test our third criterion for the desired reward shaping algorithm --- does corrected DPBA accelerate the learning of a shaped agent with good expert advice? 

\begin{figure}
\figuretitle{Corrected DPBA with good Advice}
    \centering
         \includegraphics[width=0.45\textwidth]{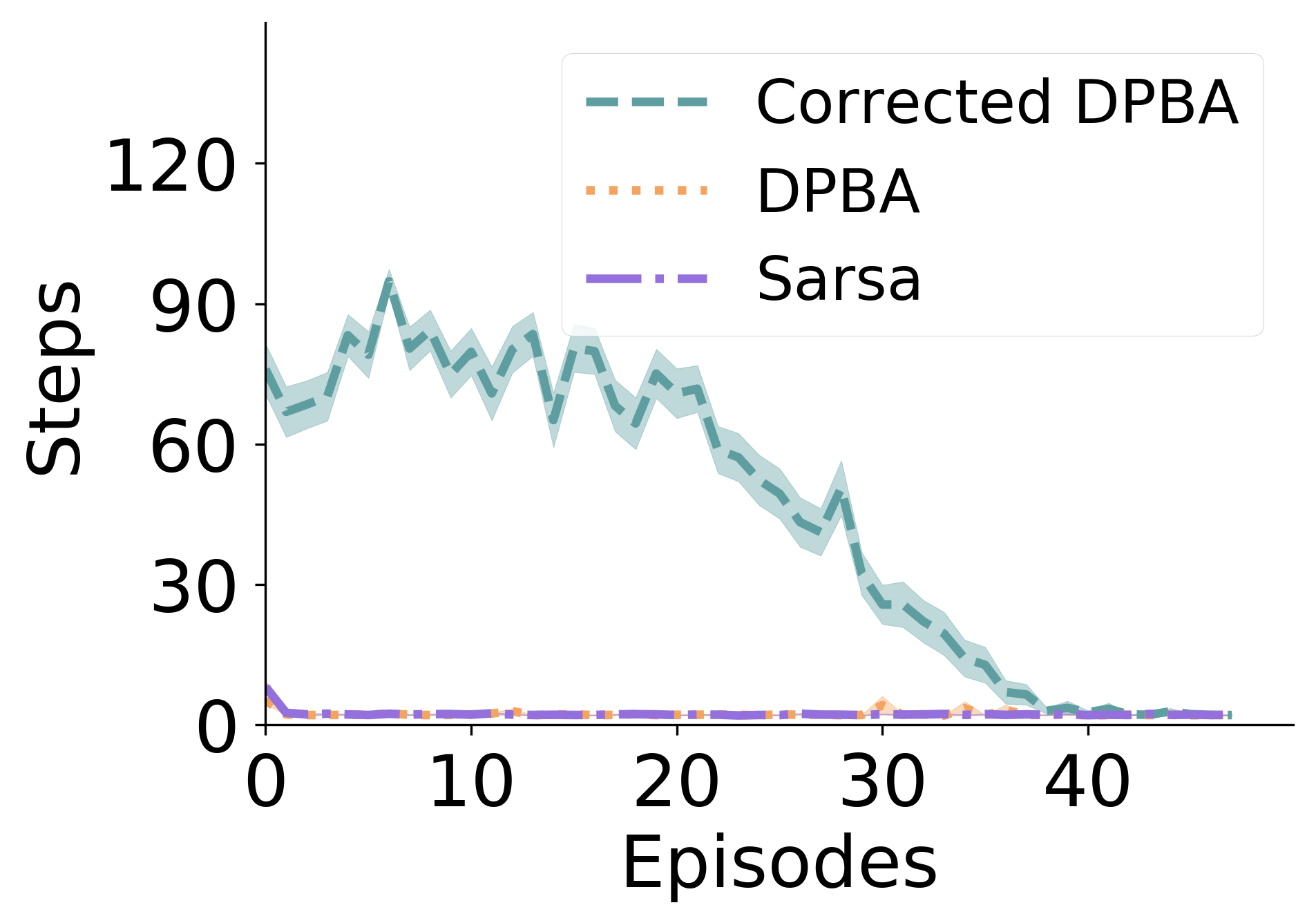}
        \caption{The y-axis shows number of time steps taken to finish each episode with the good expert. The shaped agent with corrected DPBA is compared with the shaped agent with DPBA and an unshaped Sarsa agent. Shaded areas correspond to the standard error averaged over 50 runs.}
        \label{fig:toy_example_corrected_good}
\end{figure}

Figure \ref{fig:toy_example_corrected_good} shows the results for repeating the same experiment as the previous subsection but with the good expert which is shown in Figure \ref{fig:toy_example} (b) (i.e., from each state the expert encourages the agent to move towards the goal). Here, since the expert is encouraging the agent to move towards the goal, we expect the shaped agent to learn faster than the agent that is not receiving a shaping reward. However, Figure \ref{fig:toy_example_corrected_good} shows that the corrected agent does not learn faster with good advice. To our surprise, the advice actually slowed down the learning, even though the corrected DPBA agent did eventually discover the optimal policy, as expected. 
To explain the corrected DPBA behaviour, one needs to look closely at how the $Q$ and $\Phi$ estimates are changing. The corrected DPBA adds the latest value of $\Phi$ at each time step, to correct the shaped $Q$ value; however, the $\Phi$ value which has been used earlier to shape the reward function might be different than the latest value. Let us consider the case that $\Phi$ has been initialized to zero and the advice is always a positive signal, enforcing the $\Phi$ values to be negative. With such a $\Phi$ the latest $\Phi$ values are more negative than the earlier values used for shaping the reward, which in fact discourages the desired behaviour. 

% To observe what would happen when the $\Phi$ values stabilize we can use the analyse in the original paper for the time after the convergence to the TD-fixpoint, $\Phi^{\pi}$ (Equation 19). Writing the shaping reward with the converged $\Phi$ values for an arbitrary tuple $(s,a,s',a')$ we would have:
% \begin{equation*}
%     F(s,a,s',a') = \gamma \Phi^{pi}(s',a') - \Phi^{s,a}.
% \end{equation*}
% Here the latest value of $\Phi$ which is added to correct the shaped value function, is as same as the value used to shape the reward, so it can cancel out it's peer term. However, 
% In particular, $\Phi_t$ itself is anti-correlated with good advice — it’s learned on the negative of the advice. The latest $\Phi_t$ is larger in absolute magnitude than the one used to shape $Q$ in the past, assuming that the value functions have been initialized to zero (or a near-zero value). Therefore, adding these latest $\Phi_t$ back in for selecting actions, will automatically discourage the desired behavior.
While the corrected DPBA guarantees policy invariance, it fails in satisfying the third goal for ideal reward shaping (i.e., speed up learning of an agent with a helpful advice).

The main conclusion of this section is that none of the mentioned reward shaping methods for incorporating expert advice satisfies the three ideal goals. DPBA as proposed in \citep{harutyunyan2015expressing} can lead to faster learning if the expert offers good advice but it is not policy invariant. The corrected DPBA proposed in this section is provably policy invariant but it can lead to slower learning even when provided with good advice.

\section{Policy Invariant Explicit Shaping}

In this section, we introduce the \textit{policy invariant explicit shaping }(PIES) algorithm that satisfies all of the goals we specified in Section \ref{sec:intro} for reward shaping. PIES is a simple algorithm that admits arbitrary advice, is policy invariant, and speeds up the learning of the agent depending on the expert advice. 

Unlike potential based reward shaping, the main idea behind PIES is to use the expert advice explicitly without modifying the original reward function. Not changing the reward function is the principal feature that both simplifies PIES and makes analysing how it works easier. The PIES agent learns the original value function $Q_M$ as in Equation \ref{eq:updaterule}, while concurrently learning a secondary value function $\Phi$ based on expert advice as in Equation \ref{eq:phi_update_rule}. To exploit the arbitrary advice, the agent explicitly biases the agent's policy towards the advice by adding $-\Phi$ to $Q_{M}$ at each time step $t$, weighted by a parameter $\xi_{t}$, where $\xi_t$ decays to 0 before the learning terminates\footnote{We add the negation of $\Phi$ to shape the $Q$ values since $\Phi$ is accumulating the $-R^{expert}$. We kept the $\Phi$ reward function, $R^{\Phi}$, the same as DPBA for the sake of consistency. However, one can easily revert the sign to set $R^{\Phi}=R^{expert}$ and add $\Phi$ to $Q$ to shape the agent with PIES.}. For example, for a Sarsa(0) agent equipped with PIES, when the agent wants to act greedily, it will pick the action that maximizes $Q_t(s,a)-\xi_t\Phi_t(s,a)$ at each time step. The optimal policy would be:
\begin{equation*}\label{eq:policy_soft}
\pi^{*}_{M}(s) = \arg\max_{a \in A} \left(Q^{*}_{M}(s,a) - \xi_t \Phi_t(s,a)\right),
\end{equation*}
The parameter $\xi_t$ controls that to what extent the agent's current behaviour is biased towards the advice. Decaying $\xi_t$ to 0 over time removes the effect of shaping, guaranteeing that the agent will converge to the optimal policy, making PIES policy-invariant. The speed of decaying $\xi$ determines how long the advice will continue to influence the agent's learned policy.
%The idea is to learn the original value function $Q_M$ as in \ref{eq:updaterule} without modifying the reward function, while learning the secondary value function $\Phi$ as in \ref{eq:phi_update_rule}. To exploit the arbitrary advice, instead of changing the reward function, the agent acts greedily at each time step with respect to $Q(s,a)+\xi_t\Phi_t(s,a)$, where $\xi_t$ decays from 1 to 0 before the learning terminates. So the optimal policy would be:
%\begin{equation}\label{eq:policy_soft}
%\pi^{*}_{M}(s) = \arg\max_{a \in A} \left(Q^{*}_{M}(s,a) + \xi_t \Phi_t(s,a)\right).
%\end{equation}
Choosing the decay speed of $\xi$ can be related to how beneficial it is to utilize the advice and can be done in many different ways. For this paper, we only decrease $\xi$ at the end of each episode with a linear regime. More specifically, the value of $\xi$ during episode $e$ is:
% \ys{We will find a place for this equation. I am not sure yet if it belongs here.}
\begin{equation}
    \xi_{e} := \begin{cases}
    \xi_{e-1} - \frac{1}{C} & \text{if $\xi_{e-1}$ is not } 0 \\
    0 & \text{otherwise}
    \end{cases}, \xi_1=1,
\end{equation}
where $C$ is a constant that determines how fast $\xi$ will be decayed; i.e., the greater $C$ is the slower the bias decays.

We first demonstrate empirically that PIES fulfills all three goals in the toy example.  We then show how it performs against previous methods in the grid-world and the cart-pole problems (which were originally tested for DPBA), when provided with good advice. All the domains specifications are the same as before. 
% \MET{I don't understand. Don't we just want to say we validate all three properties in both grid-world and cart-pole, with both helpful and unhelpful advice?}

\begin{figure}[ht!]
\figuretitle{PIES in Toy Example}
    \centering
    \begin{subfigure}{0.43\textwidth}
        \centering
        \includegraphics[width=\textwidth]{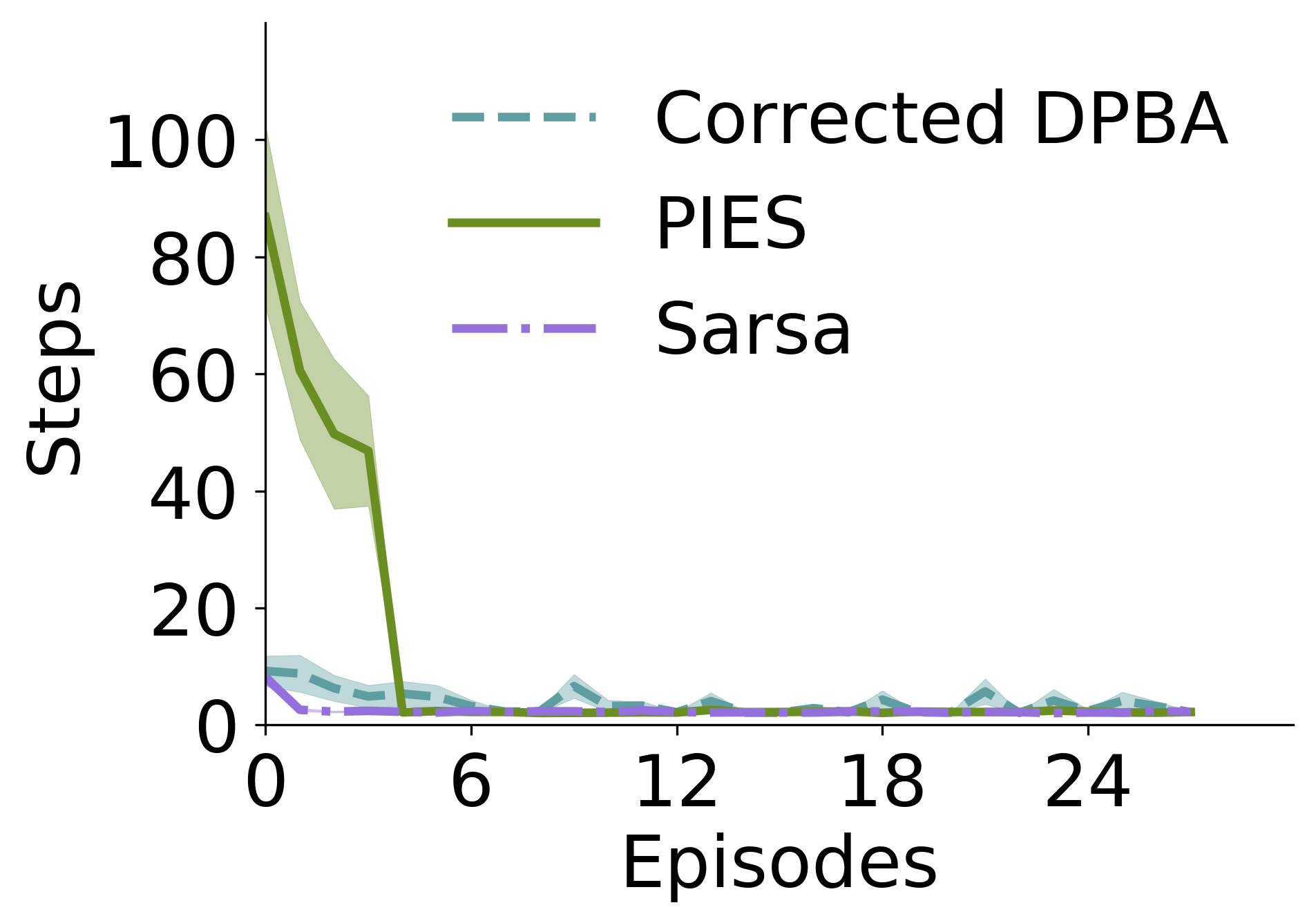}
        \caption{Advice from the bad expert}
        \label{fig:soft_smallMDP_bad}
    \end{subfigure}
    \begin{subfigure}{0.43\textwidth}
        \centering
        \includegraphics[width=\textwidth]{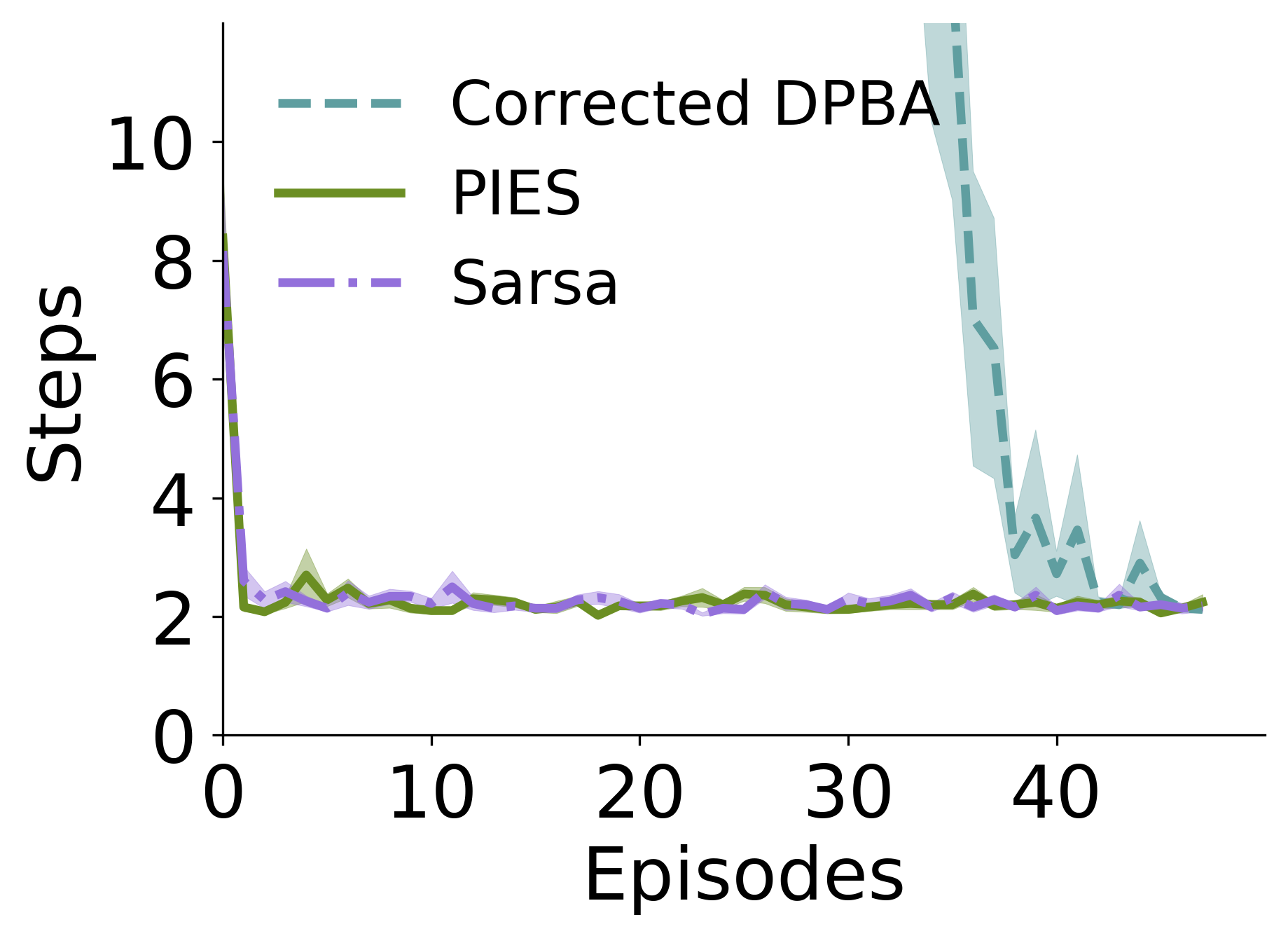}
        \caption{Advice from the good expert}
        \label{fig:soft_smallMDP_good}
    \end{subfigure}
    \caption{The y-axis shows the number of steps taken to finish each episode in the toy example. The figures compare PIES with the corrected DPBA and a Sarsa learner without shaping when the advice is a) bad and b) good. Shaded areas correspond to the standard error over 50 runs.}
    \label{fig:soft_toy_example}
\end{figure}

% \MET{All plots should have a title. Maybe just the domain used?}
The plot of the agents' performance in the toy example 
% \MET{You mean the grid-world?} 
in Figure \ref{fig:soft_toy_example} shows learning curves of the corrected DPBA, PIES, and the Sarsa learner.
% , in terms of the length of each episode
Figure \ref{fig:soft_toy_example} (a) is for the bad expert shown in \ref{fig:toy_example} (a) and Figure \ref{fig:soft_toy_example} (b) is for the good expert as in \ref{fig:toy_example} (b). Sarsa(0) was used to estimate state-action values with $\gamma=0.3$ and an $\epsilon$-greedy policy. $\Phi$ and $Q$ were initialized to 0 and $\epsilon$ decayed from 0.1 to 0. For learning rates of $Q$ and $\Phi$ value functions, $\alpha$ and $\beta$, we swept over the values $[0.05, 0.1, 0.2, 0.5]$. The values studied for setting $C$ were $[5, 10, 20, 50]$. It is worth mentioning that we set the decaying speed of $\xi$ (through setting $C$) according to the quality of the advice; i.e., a smaller $C$ for the bad advice was better as it decayed the effect of the adversarial bias faster while a larger $C$ was better with the good advice as it slowed down the decay. The best values were used, according to the AUC of each line. The learning parameters are reported in Table \ref{table:soft_toy_example}.
% were set as follows: $\alpha=0.05$, $\beta=0.2$, and $C=50$ for the good advice and $\alpha=0.1$, $\beta=0.2$, and $C=5$ for the bad advice.
% \begin{table}[ht!]
% \centering
% \caption{Parameters values for Figure \ref{fig:soft_toy_example}}
% \label{table:soft_toy_example}
% \begin{tabular}{c|ccc|}
% \cline{2-4}
%  & $\alpha$ & $\beta$ & $C$ \\ \hline
% \multicolumn{1}{|c|}{Sarsa} & 0.05 & - & - \\ \hline
% \multicolumn{1}{|c|}{corrected DPBA, good advice} & 0.2 & 0.1 & - \\ \hline
% \multicolumn{1}{|c|}{corrected DPBA, bad advice} & 0.05 & 0.2 & - \\ \hline
% \multicolumn{1}{|c|}{PIES, good advice} & 0.05 & 0.2 & 50 \\ \hline
% \multicolumn{1}{|c|}{PIES, bad advice} & 0.1 & 0.2 & 5 \\ \hline
% \end{tabular}
% \end{table}
\begin{table}[ht!]
    \caption{Parameters values for Figure \ref{fig:soft_toy_example}}
    \label{table:soft_toy_example}
  \begin{tabular}{cccl}
    \toprule
    Agent&$\alpha$&$\beta$&$C$\\
    \midrule
    Sarsa & 0.05 & - & -\\
    corrected DPBA, good advice & 0.2 & 0.1 & -\\
    corrected DPBA, bad advice & 0.05 & 0.2 & -\\
    PIES, good advice & 0.05 & 0.2 & 50\\
    PIES, bad advice & 0.1 & 0.2 & 5\\
  \bottomrule
\end{tabular}
\end{table}
% \begin{table}[ht!]
% \centering
% \caption{Parameters values for the soft-shaped agent in Figure \ref{fig:soft_toy_example}}
% \label{table:soft_toy_example}
% \begin{tabular}{c|ccc|}
% \cline{2-4}
%  & $\alpha$ & $\beta$ & $C$ \\ \hline
% \multicolumn{1}{|c|}{good advice} & 0.05 & 0.2 & 15 \\ \hline
% \multicolumn{1}{|c|}{bad advice} & 0.1 & 0.2 & 5 \\ \hline
% \end{tabular}
% \end{table}

As expected, with PIES the agent was able to find the optimal policy even with the bad expert. In the case of the beneficial advice, PIES enabled the agent to learn the task faster. The speed-up, though, is not remarkable in the toy problem, as the simple learner is also able to learn in very few episodes. 

\begin{figure}[h!]
\figuretitle{PIES in Grid-World and Cart-Pole}
    \centering
    \begin{subfigure}{0.43\textwidth}
        \centering
        \includegraphics[width=\textwidth]{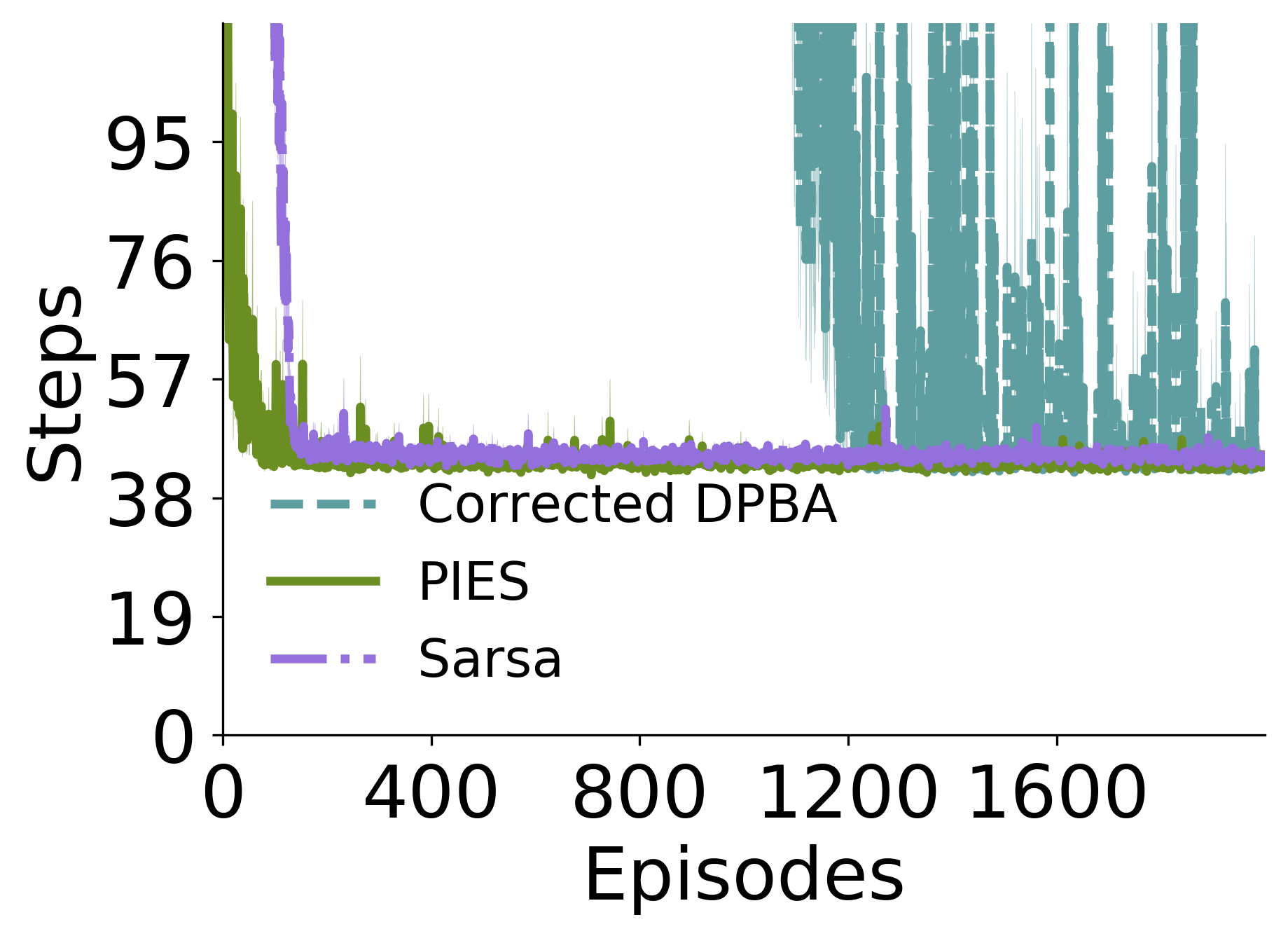}
        \caption{Grid-world}
        \label{fig:soft_gw}
    \end{subfigure}
    \begin{subfigure}{0.43\textwidth}
        \centering
\includegraphics[width=\textwidth]{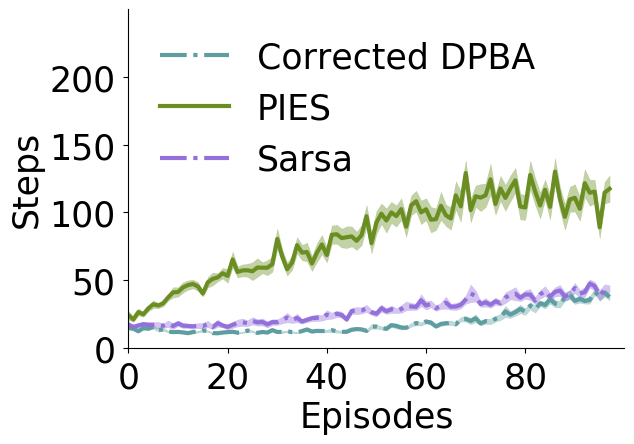}
        \caption{Cart-pole}
        \label{fig:soft_cp}
    \end{subfigure}
    \caption{The length of each episode as the number of steps in (a) the grid-world, and (b) the cart-pole domain. The plot depicts PIES versus the corrected DPBA and a Sarsa learner without shaping. Shaded areas correspond to the standard error over (a) 50 and (b) 30 runs.}
    \label{fig:soft_gw_cp}
\end{figure}

Figure \ref{fig:soft_gw_cp} better demonstrates how PIES boosts the performance of the agent learning with good advice in two more complex tasks: the grid-world and the cart-pole domains which were described in Section \ref{sec:dynamic_PBRS}. 
For both tasks similar learning parameters (those that we do not re-state) inherited their values from previous experiments. To find the best $C$, values of $[50, 100, 200, 300]$ were swept. In the grid-world task (Figure \ref{fig:soft_gw_cp} (a)) $\alpha$ and $\beta$ were selected over the values of $[0.05, 0.1, 0.2, 0.5]$. In the cart-pole task (Figure \ref{fig:soft_gw_cp} (b)), for setting the learning rates, the values of $[0.001, 0.002, 0.01, 0.1, 0.2]$ were swept. 
% The results for the grid-world task are depicted in Figure \ref{fig:soft_gw_cp} (a). For this task the learner used Sarsa(0) with $\gamma=0.99$. $\Phi$ and $Q$ were initialized to 0 and the agent selected actions according to an $\epsilon$-greedy policy with $\epsilon=0.1$. The learning rates of $Q$ and $\Phi$ value functions, $\alpha$ and $\beta$, were chosen as the best values over $[0.05, 0.1, 0.2, 0.5]$. In the cart-pole task for Figure \ref{fig:soft_gw_cp} (b), learning was done with linear function approximation via Sarsa($\lambda$) and a tile-coded feature representation. The weights for $Q$ and $\Phi$ were initialized uniformly random between 0 and 0.001. For tile-coding, we reused the parameters setting from Figure \ref{fig:pbrs} (b). $\lambda$ was set to 0.9 and $\gamma$ to 1. To choose the best learning rates of $Q$ and $\Phi$ value, $\alpha$ and $\beta$, we the values $[0.001, 0.002, 0.01, 0.1, 0.2]$ were swept over. For both tasks, the values studied for setting $C$ were $[50, 100, 200, 300]$.
As before, the best parameter values according to the AUC of each line for Figures \ref{fig:soft_gw_cp} (a) and \ref{fig:soft_gw_cp} (b), and are reported in Tables \ref{table:soft_gw} and \ref{table:soft_cp}, respectively. Just like before, for the cart-pole task's plot the upper lines indicate better performance whereas for the grid-world the lower lines are better. 

PIES correctly used the good advice in both domains and improved learning over the Sarsa learner, without changing the optimal policy (i.e PIES approached the optimal behaviour with a higher speed compared to the Sarsa learner). PIES performed better than the corrected DPBA as expected, since the corrected DPBA is not capable of accelerating the learner with good advice.
PIES is a reliable alternative for DPBA when we have an arbitrary form of advice, regardless of the quality of the advice. As shown, PIES satisfies all three desired criteria.

% \MET{Figure 5b: I'd keep the order of the algorithms consistent in the legend across different plots. Similarly, I wouldn't include a value of eta in the PIES line of the legend or have a comma after PBRS in the legend.}

% \begin{table}[ht!]
% \centering
% \caption{Parameters values for Figure \ref{fig:soft_gw}}
% \label{table:soft_gw}
% \begin{tabular}{c|ccc|}
% \cline{2-4}
%  & $\alpha$ & $\beta$ & $C$ \\ \hline
% \multicolumn{1}{|c|}{Sarsa} & 0.05 & - & - \\ \hline
% \multicolumn{1}{|c|}{corrected DPBA} & 0.1 & 0.01 & - \\ \hline
% \multicolumn{1}{|c|}{PIES} & 0.05 & 0.5 & 100 \\ \hline
% \end{tabular}
% \end{table}
\begin{table}[ht!]
  \caption{Parameters values for Figure \ref{fig:soft_gw}}
  \label{table:soft_gw}
  \begin{tabular}{cccl}
    \toprule
    Agent&$\alpha$&$\beta$&$C$\\
    \midrule
    Sarsa & 0.05 & - & -\\
    corrected DPBA & 0.1 & 0.01 & -\\
    PIES & 0.05 & 0.5 & 100\\
  \bottomrule
\end{tabular}
\end{table}
% \begin{table}[ht!]
% \centering
% \caption{Parameters values for Figure \ref{fig:soft_cp}}
% \label{table:soft_cp}
% \begin{tabular}{c|ccc|}
% \cline{2-4}
%  & $\alpha$ & $\beta$ & $C$ \\ \hline
% \multicolumn{1}{|c|}{Sarsa} & 0.1 & - & - \\ \hline
% \multicolumn{1}{|c|}{corrected DPBA} & 0.02 & 0.1 & - \\ \hline
% \multicolumn{1}{|c|}{PIES} & 0.2 & 0.5 & 200 \\ \hline
% \end{tabular}
% \end{table}
\begin{table}[ht!]
  \caption{Parameters values for Figure \ref{fig:soft_cp}}
  \label{table:soft_cp}
  \begin{tabular}{cccl}
    \toprule
    Agent&$\alpha$&$\beta$&$C$\\
    \midrule
    Sarsa & 0.1 & - & -\\
    corrected DPBA & 0.02 & 0.1 & -\\
    PIES & 0.2 & 0.5 & 200\\
  \bottomrule
\end{tabular}
\end{table}

\section{Related Work}\label{related_work}
PIES is designed to overcome the flaws we exposed with DBPA and thus tackles the same problem as DPBA. Hence, the main contrast of PIES with methods such as TAMER~\cite{knox2012reinforcement} and heuristically accelerated Q-learning (HAQL)~\cite{bianchi2004heuristically} is similar to that of DPBA with these methods: their primary difference is in how advice is incorporated. PIES and DBPA use external advice by learning a \emph{value function} online. TAMER, on the other hand, fits a model to the advice in a supervised fashion, and in some of its variants, it uses the learned model of the immediate expert reward to bias the policy. Similar to TAMER, HAQL biases the policy with a heuristic function, and is identical to the mentioned TAMER variant, if the heuristic function is substituted with a learned model of the advice. 
While TAMER and HAQL do not restrict the form of external advice, we argue that incorporating the advice in form of a value function as PIES does is more general. This is similar to the case of standard RL, where we work to maximize total discounted future rewards instead of acting myopically only based on the immediate reward. Unlike other methods, PIES emphasizes the role of the decay factor which makes PIES policy invariant. With PIES, the agent explicitly reasons about long-term consequences of following external advice, and that successfully accelerates learning, particularly in the initial (and most critical) steps.

Another line of work, related to multi-objectivization of RL by reward shaping~\cite{brys2014multi,brys2017multi}, uses multiple potential functions through PBRS. This research is orthogonal to PIES, but could be combined with it if multiple sources of advice are present.

Other approaches  \cite{gullapalli1992shaping,grezes2010online,marom2018belief} share some commonality with PIES and DBPA in that they try to approximate a good potential function. However, they do not tackle the challenge of incorporating arbitrary external advice.

\section{Conclusion}
This paper exposed a flaw in DPBA, a previously published algorithm with the aim of shaping the reward function with an arbitrary advice without changing the optimal policy. We used empirical and theoretical arguments to show that it is not policy invariant, a key criterion for reward shaping. 
Further, we derived the corrected DPBA algorithm with a corrected bias component. However, based on our empirical results the corrected algorithm fails to improve learning when leveraging useful advice resulting in a failure to satisfy the speed-up criterion. To overcome these problems, we proposed a simple approach, called PIES. We show theoretically and empirically that it guarantees the convergence to the optimal policy for the original MDP, agnostic to the quality of the arbitrary advice while it successfully speeds up learning from a good advice. Therefore PIES satisfies all of the goals for ideal shaping.

\bibliography{ref} 
\bibliographystyle{ACM-Reference-Format}
\end{document}